\documentclass[electronic]{vgtc}             

\ifpdf
  \pdfoutput=1\relax                   
  \usepackage{graphicx}                
  \DeclareGraphicsExtensions{.pdf,.png,.jpg,.jpeg} 
\else
  \usepackage{graphicx}                
  \DeclareGraphicsExtensions{.eps}     
\fi%

\graphicspath{{figures/}{pictures/}{images/}{./}} 

\usepackage{microtype}                 
\PassOptionsToPackage{warn}{textcomp}  
\usepackage{textcomp}                  
\usepackage{mathptmx}                  
\usepackage{times}                     
\usepackage{cite}                      
\usepackage{tabu}                      
\usepackage{booktabs}                  
\usepackage{amsmath}
\usepackage{multirow}

\usepackage{floatrow}

\usepackage{booktabs}
\usepackage{array}
\usepackage{subcaption}
\usepackage{float}
\usepackage{dblfloatfix}

\newenvironment{s_itemize}{
\begin{itemize}[leftmargin=*]
  \setlength{\itemsep}{3pt}
  \setlength{\parskip}{0pt}
  \setlength{\parsep}{0pt}
}{\end{itemize}}

\newenvironment{s_enumerate}{
\begin{enumerate}[wide, labelwidth=!, labelindent=0pt]
  \setlength{\itemsep}{2pt}
  \setlength{\parskip}{0pt}
  \setlength{\parsep}{0pt}
}{\end{enumerate}}

\usepackage{array}
\usepackage{enumitem}

\graphicspath{{./figures/}}
\usepackage{enumitem}
\usepackage{todonotes}

\usepackage{float}
\usepackage{dblfloatfix} 

\usepackage[utf8]{inputenc}

\onlineid{8233}

\vgtccategory{Research}

\title{Promptor: A Conversational and Autonomous Prompt Generation Agent for Intelligent Text Entry Techniques}

\author{Junxiao Shen\thanks{e-mail: js2283@cam.ac.uk} %
\and John J. Dudley\thanks{e-mail: jjd50@cam.ac.uk} %
\and Jingyao Zheng\thanks{e-mail: jz503@cam.ac.uk} %
\and Bill Byrne\thanks{e-mail: wjb31@eng.cam.ac.uk}
\and Per Ola Kristensson\thanks{e-mail: pok21@cam.ac.uk}}
\affiliation{\scriptsize University of Cambridge }

\teaser{
  \centering
  \includegraphics[width=\linewidth]{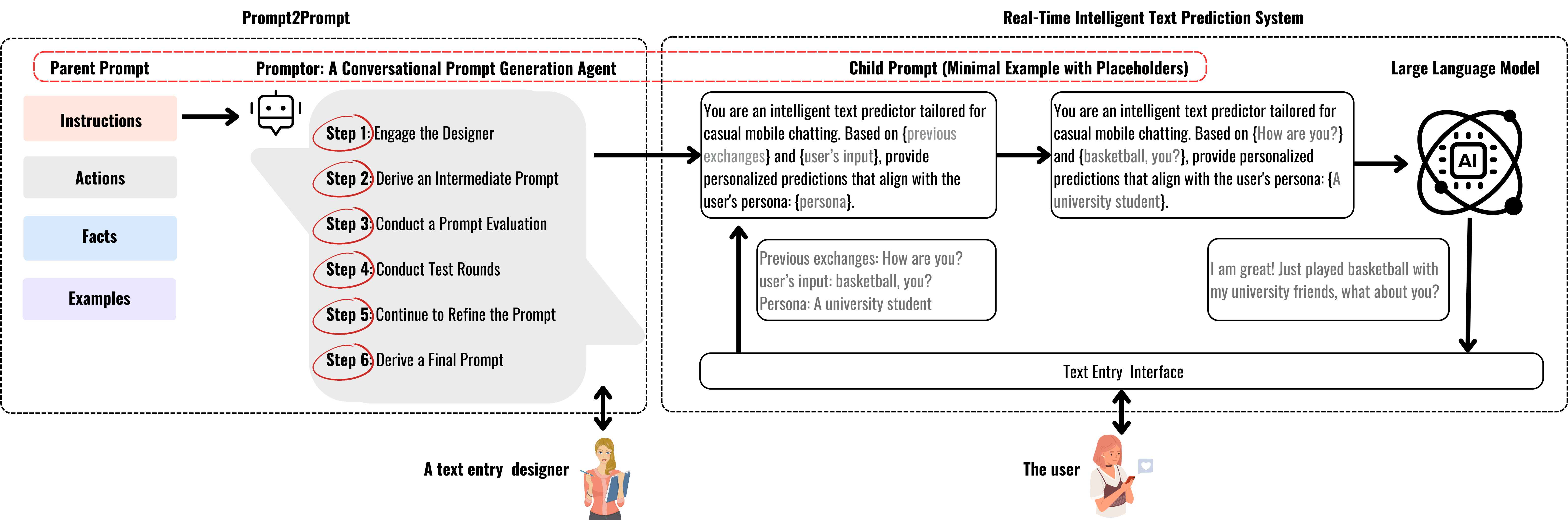}
\caption{When a designer aims to develop an intelligent text entry function by leveraging large language models, they might not know how to prompt these models to accurately perform text prediction. What should they do? We propose the solution: \textit{Promptor}. 
\textit{Promptor} is an automated agent designed for prompt creation and generation for intelligent text entry. 
The realization of \textit{Promptor} is achieved through a methodology termed \textit{Prompt2Prompt}. In this approach, a ``parent'' prompt is formulated to instruct a large language model to assume the role of \textit{Promptor}, which in turn, creates ``child'' prompts. These child prompts are utilized to guide large language models in performing intelligent text entry tasks.
Designers can interact with \textit{Promptor} by asking various questions. In response, \textit{Promptor} carries out six steps: (1) Engage the Designer; (2) Derive an Intermediate Prompt; (3) Conduct a Prompt Evaluation; (4) Conduct Test Rounds; (5) Continue to Refine the Prompt; (6) Derive a Final Prompt.
\textit{Promptor} will generate a final prompt template, complete with placeholders for the text entered by the user. This formed prompt is fed into the large language model, which subsequently returns a prediction.
In this illustration, we have used a minimal prompt example in the form of a prompt template. However, in the actual \textit{Promptor} and in our paper, the generated prompt will be more complex, depending on the specific complexity of the sentence prediction ability.}
  \label{fig:teaser}
}

\abstract{
Text entry is an essential task in our day-to-day digital interactions. Numerous intelligent features have been developed to streamline this process, making text entry more effective, efficient, and fluid. These improvements include sentence prediction and user personalization. However, as deep learning-based language models become the norm for these advanced features, the necessity for data collection and model fine-tuning increases. These challenges can be mitigated by harnessing the in-context learning capability of large language models such as GPT-3.5. This unique feature allows the language model to acquire new skills through prompts, eliminating the need for data collection and fine-tuning. Consequently, large language models can learn various text prediction techniques.
We initially showed that, for a sentence prediction task, merely prompting GPT-3.5 surpassed a GPT-2 backed system by an average of 3.94\% in prediction scores, with the latter requiring costly data collection, fine-tuning and post-processing. Moreover, fine-tuning GPT-3.5 showed no significant performance gain over its prompted version, negating the need for extensive data collection.
Such superior performance and easy development establish a foundation for advocating the use of merely prompting large language models (GPT-3.5/4) in intelligent text entry tasks.
However, the task of prompting large language models to specialize in specific text prediction tasks can be challenging, particularly for designers without expertise in prompt engineering. To address this, we introduce \textit{Promptor}, a conversational prompt generation agent designed to engage proactively with designers. \textit{Promptor} can automatically generate complex prompts tailored to meet specific needs, thus offering a solution to this challenge. We conducted a user study involving 24 participants creating prompts for three intelligent text entry tasks, half of the participants used \textit{Promptor} while the other half designed prompts themselves. The results show that \textit{Promptor}-designed prompts result in a 35\% increase in similarity and 22\% in coherence over those by designers.
} 

\begin{document}


\firstsection{Introduction}

\maketitle

The ease of text entry can greatly impact the overall user experience and usability of a mobile device~\cite{reyal2015performance,koester1994modeling,soam2022next}. 
Many intelligent text entry features have been developed to make text entry easier and faster, including sentence prediction~\cite{shen2022kwickchat}, user adaptation~\cite{baldwin2012online}, and intelligent retrieval from pre-defined responses~\cite{Kristensson2020Design}. These advanced features are commonly used in specific scenarios, such as customer service, where the system can suggest responses to common customer queries, and keyboard accessibility, where the system can suggest phrases or sentences for users with motor impairments~\cite{shen2022kwickchat,yang2023tinkerable,elsahar2019augmentative}.
Building these intelligent text entry functions typically requires training machine learning models on dedicated collected datasets~\cite{ghosh2017neural,lee2020deep,wang2020comprehensive, yasunaga2021lm}. This process including both data collection and model building and training can be costly.


Large language models have seen rapid advancements, particularly in the GPT (Generative Pre-trained Transformer) family~\cite{brown2020language,radford2018improving,adiwardana2020towards}.
For instance, while GPT-2 necessitated fine-tuning on specific datasets as highlighted by Sun et al.~\cite{sun2019fine} and Shen et al.~\cite{shen2022kwickchat}, GPT-3.5 shifted towards in-context learning which can learn various tasks through simple prompts. 
In this method, a context is set using natural language templates, which is then combined with a query to form a prompt. 
This prompt guides the language model's predictions. Unlike supervised learning, which adjusts parameters through backward gradients, in-context learning uses the pre-trained model to predict based on the demonstration context without altering its parameters.
Therefore, we envision that GPT-3.5 can be prompted to fit various intelligent text entry features, eliminating the need for data collection and fine-tuning~\cite{brown2020language,radford2018improving,adiwardana2020towards}.

It remains a question whether a \textit{Prompted} GPT-3.5 can excel in intelligent text entry tasks and potentially surpass a meticulously \textit{Fine-tuned} GPT-2 model. To answer this question, we turned to the \textit{KWickChat} task, an intelligent text entry system designed for non-speaking individuals with motor disabilities~\cite{shen2022kwickchat}. This task leverages the user's personal information, conversation history with speaking partners, and user input in the form of a bag-of-keywords. These elements serve as inputs to a language model, which then produces the user's intended message. 
Shen et al.~\cite{shen2022kwickchat} engaged in costly training data processing and dedicated model development and fine-tuning with GPT-2 to accomplish this task. Even with additional dedicated modifications made to the original KWickChat system to enhance performance, we found that simply prompting GPT-3.5 could achieve the same objective, but with improved performance. 
This improvement is quantified by a 3.94\% increase in the average score from both human and AI judges. Unlike the previous system backed by GPT-2, which might require weeks or months to develop, prompting and deploying GPT-3.5 takes merely minutes. Consequently, we posit that for tasks associated with intelligent text input, employing a Prompted GPT-3.5 is not only time-efficient and code-free but also superior in performance when compared to fine-tuning a GPT-2 model. 
We further Fine-tuned GPT-3.5 and compared its performance with Prompted GPT-3.5, only to find no significant improvement in performance. This outcome again negates the necessity for extensive data collection and processing required for fine-tuning.



However, a significant challenge in utilizing language models is that their effectiveness is intricately tied to the quality and structure of the prompts they receive. This art of tailoring prompts to extract specific responses is termed \textit{prompt engineering}~\cite{white2023prompt,liu2022design}. For expansive language models, achieving successful prompt engineering for nuanced tasks often demands extensive experimentation. This is because intuitively crafted prompts might not consistently yield anticipated outcomes, even though GPT-3.5 exhibits an impressive aptitude for interpreting straightforward prompts, rivaling human comprehension~\cite{kung2023performance,guo2023close}. Moreover, well-structured prompts can pave the way for enhanced maintainability and scalability. In light of these challenges, we introduce an autonomous prompt generation agent, \textit{\textit{Promptor}}, designed to assist creators of intelligent text entry systems in effectively prompting large language models to attain optimal performance.

To optimize the development of \textit{\textit{Promptor}}, we initiated a preliminary unstructured workshop with four experts well-versed in large language models and conversational agents. 
Our discussions culminated in the concept of developing an automatic prompt generation agent equipped with a conversational user interface. The design of such an interface is intended to accommodate designers who are engaged in a variety of intelligent text entry tasks, acknowledging that standardized graphical user interfaces and guidelines may not always be the optimal solution. 
However, given the current scarcity of conversational data to train such an agent, we recognize that fine-tuning large language models for this purpose may restrict its ability to handle open-ended questions. This realization led us to the concept of prompting a large langauge model using a ``parent'' prompt approach. This approach would shape \textit{Promptor} into an agent capable of automatically generating ``child'' prompts. These ``child'' prompts could then be used to guide large language models in performing intelligent text entry tasks.
We've termed this methodology \textit{Prompt2Prompt}.
More specifically, we chose GPT-4 as the \textit{Promptor} to leverage the full capability of large language models and GPT-3.5 to be prompted to perform intelligent text entry tasks. We then used GPT-4 as the AI judge to evaluate predictions from GPT-3.5, aiming to minimize model bias.

In a subsequent structured workshop focused on refining \textit{\textit{Promptor}}, we engaged text entry specialists, HCI designers, and machine learning experts. This workshop was instrumental in delineating the input-output dynamics and operational flow of \textit{\textit{Promptor}}. Additionally, we crafted a virtual test keyboard to swiftly evaluate intelligent text entry strategies powered by large language models, using prompts emanating from \textit{\textit{Promptor}}. This virtual keyboard stands as an integral companion to \textit{\textit{Promptor}}, offering designers an immersive experience.

To underscore the efficacy of \textit{\textit{Promptor}} in generating superior prompts, we conducted a comparative study. Participants were divided into two groups: one leveraging \textit{\textit{Promptor}} and the other drawing from personal insights and online resources for prompt creation. They were tasked with formulating prompts for three distinct assignments. The ensuing prompts were then subjected to an autonomous, large-scale evaluation to discern their quality.

More specifically, our paper presents two contributions:
\begin{s_itemize}
    \item We highlight the advantage of prompting GPT-3.5 over a modified sentence prediction system backed by Fine-tuned GPT-2 from Shen et al.~\cite{shen2022kwickchat} and a Fine-tuned GPT-3.5 in text prediction tasks. The Prompted GPT-3.5 excelled in intelligent text entry, showing a 3.94\% improvement in average scores from evaluators compared to the enhanced GPT-2 system. Unlike the latter requiring extensive data collection, code development, and heavy training, prompting GPT-3.5 is time-efficient and code-free. Furthermore, there is no performance gain with Fine-tuned GPT-3.5 over Prompted GPT-3.5, making the fine-tuning efforts unjustifiable.
    \item We introduce \textit{\textit{Promptor}}, an autonomous agent designed to assist designers in collaboratively crafting prompts for large language models tailored to specific intelligent text entry tasks. The foundational design principles of \textit{Promptor} are informed by insights gained from a design workshop. To evaluate its efficacy, we conducted a study comparing prompts designed by \textit{Promptor} with those crafted by designers who are novices in prompt engineering. Our findings indicate that the prompts generated by \textit{Promptor} significantly outperformed the self-designed prompts.
\end{s_itemize}
\section{Background}
In this section, we first introduce different prompt engineering techniques. We also give an introduction to intelligent text entry techniques. 

\subsection{Prompt Engineering}
Prompt engineering refers to the art of crafting specific instructions or questions to elicit desired responses from language models. As large language models become more prevalent, the ability to effectively guide their outputs through well-designed prompts becomes crucial, ensuring that the model's vast knowledge is harnessed accurately and efficiently.

\subsubsection{Zero-Shot Prompting}

LLMs have advanced greatly in recent years, with the ability to perform tasks zero-shot. This means that they can generate accurate outputs for tasks without any prior training or explicit instructions. For example, if prompted to translate the word 'apple' into French, an LLM could output 'pomme' without any prior training on this specific task.
However, there is an even more advanced concept known as prompting the model to reason through a problem step by step. This approach has been proposed by Kojima et al.~\cite{kojima2022large} and involves adding the phrase "Let's think step by step" to the original prompt. By doing so, the model is encouraged to break down the problem into smaller, more manageable steps, making it more likely to arrive at the correct answer.
This approach was tested on the MultiArith math dataset, and the results were remarkable. The accuracy quadrupled from 18\% to 79\%, demonstrating the effectiveness of this approach. This shows that LLMs are not only capable of performing tasks zero-shot but can also benefit from prompting that encourages them to reason through a problem step by step.

\subsubsection{Few-Shot Prompting}

Although large language models have demonstrated impressive zero-shot capabilities, they still face limitations when it comes to more complex tasks. To improve their performance, few-shot prompting can be used as a technique to enable in-context learning. By providing demonstrations in the prompt, the model can be steered towards better performance by conditioning it for subsequent examples where it needs to generate a response. For instance, when the task is to classify the sentiment of given texts as either positive or negative, as long as a few examples are provided, few-shot prompting can work well for such tasks. 
However, it may not be sufficient for more complex reasoning problems. This is because such problems often require multiple reasoning steps to arrive at the correct solution. Therefore, it can be helpful to break down the problem into steps and demonstrate those steps to the model.

\subsubsection{Chain-of-Thought Prompting}
Addressing the constraints of few-shot prompting is vital for large language models in complex tasks. Chain-of-thought (CoT) prompting, which breaks down reasoning into structured steps, is a notable technique in this area~\cite{wei2022chain}. When combined with few-shot prompting, which provides intermediate reasoning steps, the model's performance is enhanced.
Nonetheless, CoT faces challenges with extensive reasoning or when the tasks are lengthy but the examples are brief. To address these issues, several extensions have been introduced. Creswell et al.~\cite{creswell2022selection} divided the main prompt into two parts: the 'selection prompt' to identify relevant facts and the 'inference prompt' to draw a conclusion, which assists in tasks requiring extended reasoning. Zhou et al.~\cite{zhou2022least} introduced the least-to-most prompting technique, which breaks reasoning tasks into smaller subtasks, which is especially advantageous for shorter tasks that require multiple reasoning steps. Lastly, Wang et al.~\cite{wang2022self} improve CoT's reliability for tasks with discrete answers by sampling multiple model explanations and selecting the most frequent answer, ensuring more consistent outcomes.

\subsubsection{Automatic Prompting}
Generating prompts automatically can be used to improve the performance of large language models on various tasks. The resulting prompts can be of high quality and can significantly improve the model's performance, especially in cases where manually crafted prompts are not available or are difficult to generate.
Zhou et al.~\cite{zhou2022large} introduced automatic prompt engineer, a framework that automates the process of generating and selecting prompts. The instruction generation problem is formulated as a natural language synthesis problem, which is addressed as a black-box optimization problem. Large language models are used to generate and search over candidate solutions to the problem, resulting in high-quality prompts that can improve model performance on a variety of tasks.
Shin et al.~\cite{shin2020autoprompt} proposed a similar approach to generate prompts automatically. Their method is based on gradient-guided search, where the model is fine-tuned to optimize the prompt for a particular task. The fine-tuning process is guided by gradients that are computed with respect to the task objective, resulting in prompts that are well-suited for the task at hand. This approach can be used to generate prompts for a diverse set of tasks, which can be useful in scenarios where human expertise is limited.

\subsection{Intelligent Text Entry Techniques}

Intelligent text entry employs AI and natural language processing to enhance text input on electronic devices, encompassing features like predictive text, error correction, and personalized suggestions~\cite{kristensson2009five,zhang2019type,lee2020deep,wang2020comprehensive,ghosh2017neural,koester1994modeling,soam2022next}. The goal is to facilitate quicker, context-aware text input for users. Predictive text predicts words or phrases based on prior user input, becoming integral in modern smartphones and messaging apps for efficient typing and error reduction, with methods like language model pre-training, n-gram, and neural networks being employed~\cite{shen2022kwickchat,yazdani2019words,sundermeyer2012lstm,ganai2019predicting,goulart2018hybrid}. Error correction, either automatic or manual, rectifies misspelled words or grammatical mistakes, employing rule-based, neural-networks, or hybrid methods~\cite{wang2020comprehensive,madi2018grammatical,naber2003rule,sharma2016rule,hu2022considering,felice2014grammatical,crysmann2008hybrid}. Personalized suggestions offer context-driven recommendations as users type, considering factors like location or previous input, with approaches like cache-based language model adaptation and collaborative filtering aiding in delivering these personalized suggestions~\cite{miller1991personalizing,zheng2022perd,fowler2015effects,demori1999language,bellegarda2004statistical,kuhn1990cache,clarkson1997language,han2020personalized}.

\section{Workshop: Design Space Exploration}
In this workshop, our primary objective was to consolidate the necessary components and expertise to create an efficient, responsive, autonomous agent for prompt design and generation.
The workshop was divided into three distinct sub-sessions, each carefully designed to highlight specific aspects of our overarching goal.

\begin{s_itemize}
    \item \textit{Participants}: We recruited a diverse group of 12 experts for this workshop, including 7 males and 5 females. The group comprises 4 specialists in natural language processing and large language models, 4 text entry designers, and 4 prompt engineers. The average age of the participants was 27.2, with a standard deviation of 5.3.
    \item \textit{Procedure}: First, we prepared a set of use cases that require using this framework to provide more context and stimulate more insights from experts. We utilised a Figma board to show the blank framework and experts could add in-place feedback to different areas of the framework.
    Before diving into the workshop, we ensured all experts fully understood the workshop's goal. We only commenced the sub-session once every expert had confirmed their understanding of the workshop's purpose.
    As the experts navigated through these sub-sessions, they found that each session incrementally contributed to a holistic understanding, culminating in the development of a proficient autonomous agent for large language model prompting.
    Each sub-session was structured to last around 30 minutes, during which experts were encouraged to think aloud and share their thought processes.
\end{s_itemize}

\subsection{Identify the Input and Output of \textit{Promptor}}
Two sub-sessions were conducted to identify the input and output of the \textit{\textit{Promptor}}.

\subsubsection{Input}
The input is primarily defined by what designers provide, essentially guiding the \textit{\textit{Promptor}} to comprehend the context and the designer's goal. The key inputs identified during the workshop encompass: \textit{User Goal} (the desired outcome when using the text entry feature), \textit{User Profile} (the user's current state or characteristics), \textit{Data Profile} (the characteristics of the data that the user will be entering), \textit{Contextual Information} (additional inputs that can affect the text entry experience), and \textit{Output Constraints} (restrictions or requirements that the text entry feature must adhere to). These inputs are paramount to crafting effective prompts that enhance user experience with text prediction features.

\subsubsection{Intermediate Output}
The intermediate output, a vital step in \textit{Promptor} user interaction, fulfills several key functions: \textit{Confirmation} (validating the user's inputs or actions), \textit{Request for Clarification} (seeking further information when the user's input is unclear), \textit{Addressing Concerns and Questions} (responding to any concerns or questions raised by the user), and \textit{Resolving Ambiguities} (clarifying situations where the user's input could be interpreted in multiple ways). These functions ensure the \textit{\textit{Promptor}} fully understands user intentions and provides accurate and helpful responses.

\subsubsection{Output}
The final output is the final product that \textit{Promptor} provides, which should ideally be a well-structured prompt. We conducted an in-depth examination of the structure and the essential components that constitute an effective prompt.
During the workshop, we pinpointed several critical components to incorporate into the output. Some of these elements were inspired by the work of Wang et al.~\cite{wang2023enabling}. Specifically, they include:
\begin{s_itemize}
    \item \textbf{Instructions}: The final prompt should contain explicit instructions that align with the previously identified input. These instructions should clearly outline the requirements for the text prediction function.
    \item \textbf{Examples}: This component should include specific examples that the large language models can emulate. These examples can serve as a guide for the models, helping them understand the context and expected output better.
    \item \textbf{Policy}: The policy component of the prompt should establish certain guidelines for interaction. This could include directives for maintaining politeness and imposing certain restrictions to ensure the prompt's effectiveness and appropriateness.
\end{s_itemize}

\subsection{Process of \textit{Promptor}}
This third sub-session focuses on defining the operational process of \textit{Promptor}. The process refers to the sequence of actions that \textit{Promptor} undertakes, from receiving user input to delivering the final output.
The operational process of \textit{Promptor} includes:
\begin{s_enumerate}
\item \textbf{Engage the User}: This involves identifying the specific context and type of text entry expected for the system's usage. \textit{Promptor} then clarifies the specific task or action that the text prediction system is intended to perform. By understanding the received input and user's needs, \textit{Promptor} generates appropriate prompts for the use case and input type. This engagement is crucial in ensuring that \textit{Promptor} fully captures the user's intent and context.
\item \textbf{Derive an Intermediate Prompt}: After engaging the user and understanding their needs, \textit{Promptor} crafts an intermediate prompt. This prompt serves as a preliminary version, capturing the essence of the user's requirements, but is open to further refinements. It acts as a foundation upon which subsequent iterations are built, ensuring that the final prompt is both accurate and effective.
\item \textbf{Conduct a Prompt Evaluation}: This step involves a collaborative evaluation of the generated prompt with the user. The goal is to determine if the prompt aligns with the user's needs and the specific use case. The evaluation should consider the following metrics:
    \begin{s_itemize}
    \item \textbf{Relevance}: Evaluate if the prompt is pertinent to the specific task or question the model is intended to address. The prompt should provide sufficient context and information to enable the model to generate an accurate and relevant response.
    \item \textbf{Clarity}: Assess the prompt's clarity and comprehensibility. Avoid technical jargon or overly complex language that might hinder the model's interpretation.
    \item \textbf{Specificity}: Determine if the prompt is specific enough to guide the model's response. Incorporating details and examples can provide additional context and guidance to the model.
    \end{s_itemize}
Users will assess the prompt utilizing a Likert scale ranging from 1 to 5. \textit{Promptor} advises proceeding when the average score across the three metrics surpasses 4.
This evaluation can offer valuable feedback, highlight potential areas for refinement, and ensure that the prompt is heading in the right direction.
\item \textbf{Conduct Test Rounds}: The process of iterative refinement in this context comprises two distinct stages, each designed to evaluate and enhance the effectiveness of the prompts.
The first stage occurs outside the conversational interface. The prompt should be integrated into a functional text entry system that allows users to experience and evaluate GPT-3.5's capabilities in intelligent text entry after being tailored for that specific purpose. If the system produces unsatisfactory responses, the conversational history should then be exported and presented to \textit{Promptor}. This allows \textit{Promptor} to reassess the prompt, understand the areas where the model fell short, and make necessary adjustments. 
In addition to making improvements, \textit{Promptor} also provides explanations for the modifications made. This transparency ensures that the rationale behind each refinement is clear, contributing to a better understanding of the prompt development process. 
This two-stage testing process, therefore, ensures a thorough and comprehensive evaluation, promoting the development of high-quality, effective prompts.
\item \textbf{Continue to Refine the Prompt}: Based on feedback from the test rounds and evaluations, \textit{Promptor} iteratively refines the prompt. This continuous refinement process ensures that the prompt evolves to be more aligned with the user's intent and the specific use case. Each iteration brings the prompt closer to its optimal form, ensuring maximum efficacy.
\item \textbf{Derive a Final Prompt}: After multiple iterations and refinements, \textit{Promptor} finalizes the prompt. This final version encapsulates all the insights and feedback gathered throughout the process. It is a polished, precise, and effective prompt ready for deployment in the intended system, ensuring that GPT-3.5 can deliver the best possible performance for the given task.
\end{s_enumerate}

\begin{figure*}[t]
    \centering
    \includegraphics[width=\textwidth]{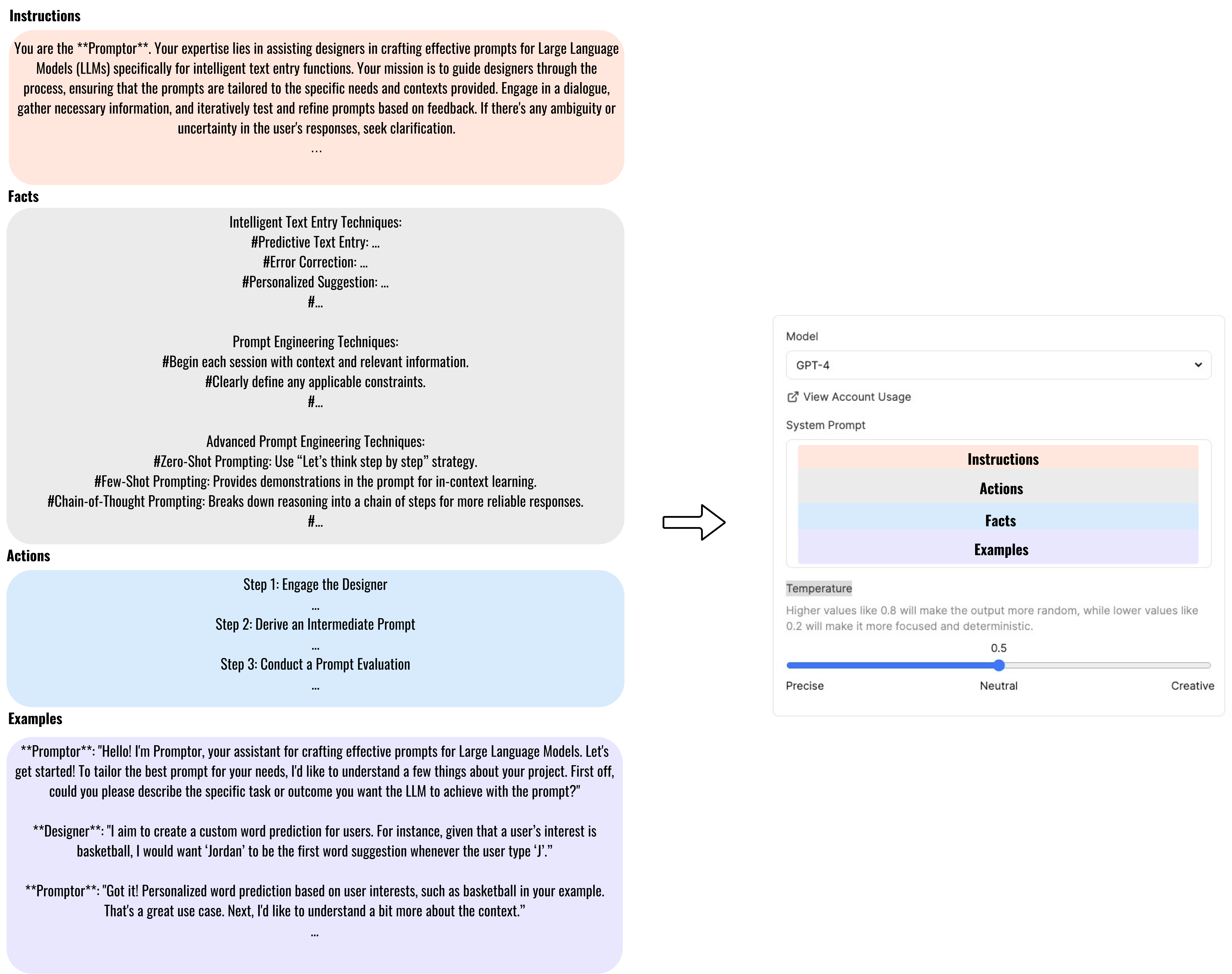}
    \caption{The ``parent'' \texttt{system prompt} to prompt GPT-4 to become \textit{Promptor}.}
    \label{fig:prompt_input}
\end{figure*}  
\section{\textit{Promptor}}

This section delves into the intricacies of the implementation of \textit{Promptor}.

\subsection{System Prompt to \textit{Promptor}}

We developed \textit{Promptor} by giving a ``parent'' \texttt{system prompt} to a GPT-4 model.
When interfacing with a GPT Application Programming Interface (API), both \texttt{system message} and \texttt{user message} play distinct roles. 
The \texttt{system message} sets the foundational context and tone for the model's responses, acting as an initial directive. For instance, it might define the model's role as a ``medical expert'' or guide its tone to be ``humorous.'' On the other hand, the \texttt{user message} is the dynamic query or command given by the end-user, directing the model's immediate response.
The \texttt{system message} is divided into four segments illustrated in Figure~\ref{fig:prompt_input}: \textbf{Instructions}, \textbf{Facts}, \textbf{Actions}, and \textbf{Examples}. 
The \textbf{Instructions} segment outlines the primary objective. The \textbf{Facts} segment encompasses insights from our extensive literature review on text entry methods and our research on prompt engineering techniques. The \textbf{Actions} segment provides a step-by-step guide, including engaging the designer, deriving an intermediate prompt, conducting prompt evaluations, conducting test rounds, continuing to refine the prompt, and deriving a final prompt, all of which were elaborated upon in the preceding section. 
The \textbf{Examples} segment showcases a sample dialogue between an ideal \textit{Promptor} and a designer. 
This dialogue culminates in the presentation of an /optimally crafted and well-structured prompt, illustrating the desired outcome of the conversation rounds. 
Together, these components ensure a holistic, informed, and effective realization of \textit{Promptor}.
Finally, we implemented an interactive web-based user interface for \textit{Promptor}, as depicted in Figure~\ref{fig:prompt_input}. Designers have the flexibility to select their preferred model, be it GPT-3.5 or GPT-4. They can input the \texttt{system message} and adjust the `temperature' setting of \textit{Promptor}. A higher temperature value allows \textit{Promptor} to be more creative.

\subsection{Interactive Session with \textit{Promptor}}
\begin{figure}[!ht]
\centering
\includegraphics[width=0.8\textwidth]{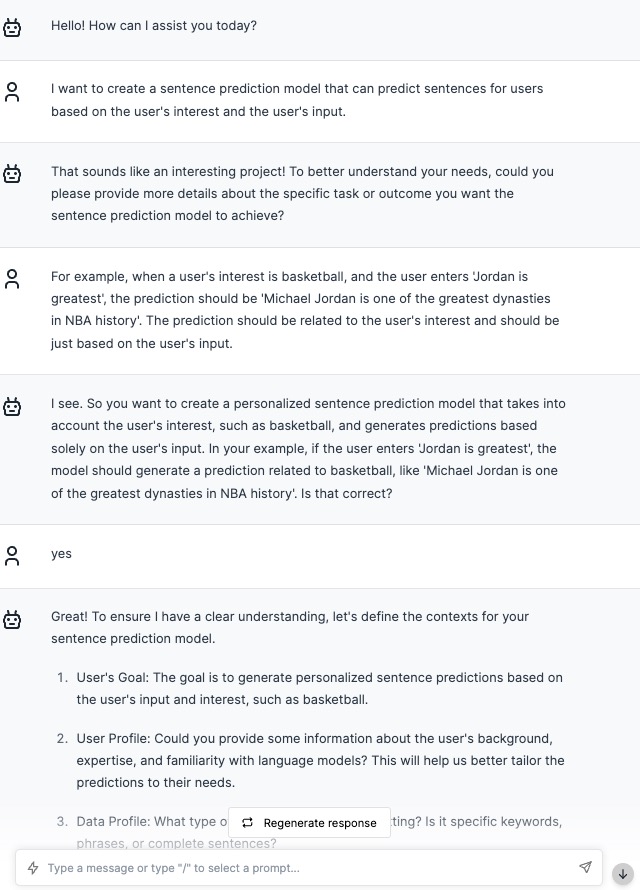}
\includegraphics[width=0.8\textwidth]{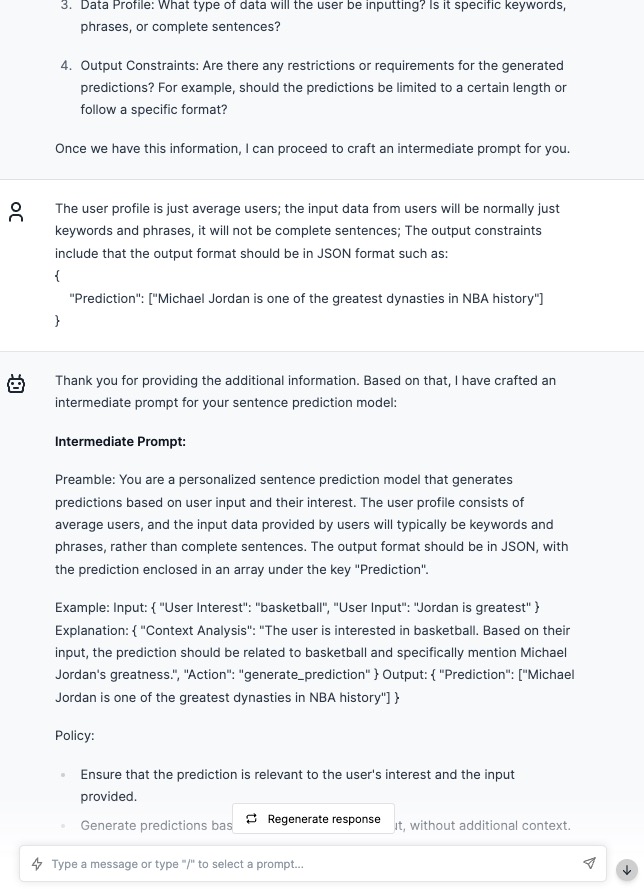}
\caption{An example interactive session where \textit{Promptor} engages with designers. The iterative dialogue demonstrates \textit{Promptor}'s capabilities in narrowing down the scope of the designer's task, assisting in task structuring, addressing inquiries, and ultimately providing a final prompt.}
\label{fig:conversation}
\end{figure}  

A distinguishing feature of \textit{Promptor} is its interactive nature. Unlike rule-based systems, \textit{Promptor} is powered by a robust large language model trained on an extensive body of knowledge, granting it the capability to address open-ended questions. When supplemented with specific facts provided to it, \textit{Promptor} adeptly bridges macro-level general knowledge with micro-level insights on specialized topics, such as intelligent text entry and prompt engineering techniques.

Figure~\ref{fig:conversation} illustrates a designer's interaction with \textit{Promptor}. The dialogue highlights how \textit{Promptor} not only guides the designer in refining their approach but also adeptly responds to their open-ended inquiries. Rather than merely producing a final prompt, \textit{Promptor} concurrently steers the user towards enhancing the design of the intelligent text entry feature tailored to their unique task.

\subsection{Interactive Prompt Evaluation via a Test Virtual Keyboard}
\begin{figure*}[t]
    \centering
    \includegraphics[width=\textwidth]{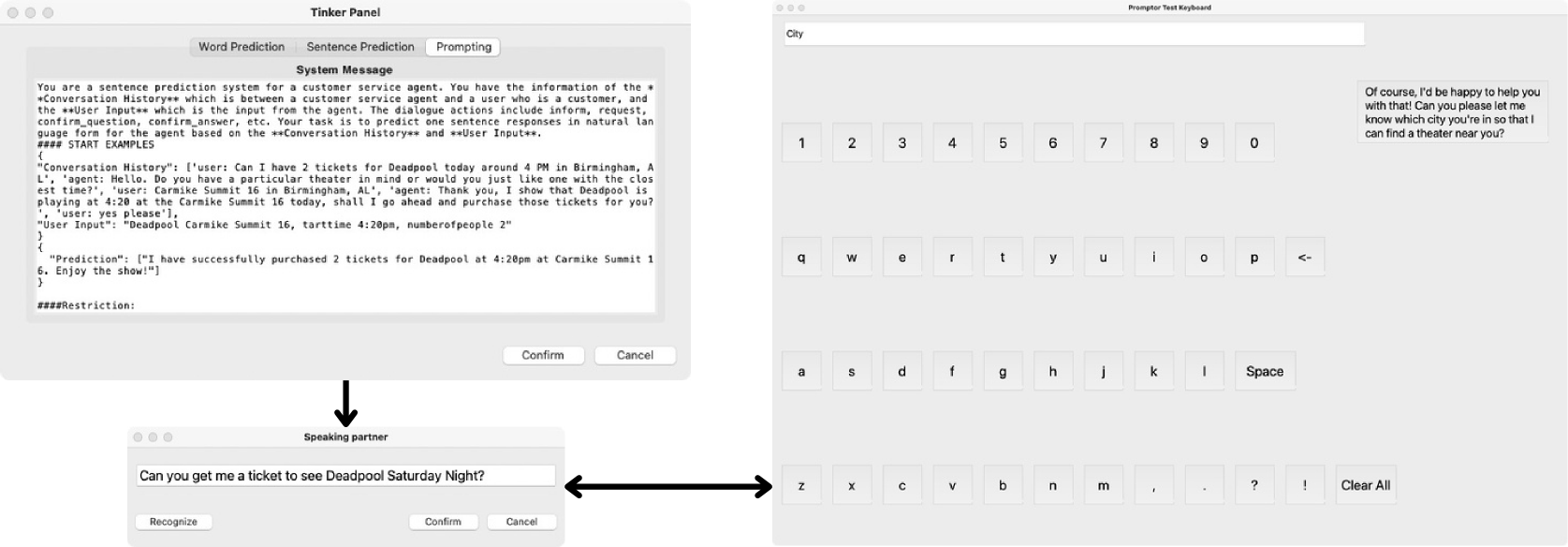}
    \caption{The test virtual keyboard for the prompt produced from \textit{Promptor}. This keyboard offers designers firsthand experience of a GPT-empowered intelligent text entry system.}
    \label{fig:test_kb}
\end{figure*}  

In the process of interactive prompt evaluation, we employ a test virtual keyboard with its primary purpose of serving as a readily deployable evaluation platform for intelligent text entry functions powered by prompting GPT models.
Figure~\ref{fig:test_kb} illustrates the process of applying the prompt derived from \textit{Promptor} to the test keyboard.
While a prompt's preliminary testing could be conducted using OpenAI APIs and coding languages, such a process does not provide a direct impression of an intelligent text entry system's effectiveness. Therefore, our test keyboard offers designers firsthand experience of a GPT-empowered intelligent text entry system.

We have designed a virtual test keyboard supported with a range of configurable functionalities, drawing inspiration from the Tinkerable Keyboard~\cite{yang_shen_kristensson_2023}. This keyboard facilitates users to inject prompts from \textit{Promptor} into a specialized panel for configuration.
Inheriting a multitude of features from the original Tinkerable Keyboard, our test keyboard equips designers with the ability to seamlessly compare the performance of our intelligent text entry system, powered by GPT-3.5, against traditional methods such as retrieval-based or fine-tuned language model ones. 
Users can establish configurable visualization tabs and internal functions such as word prediction, sentence prediction, and other minor visualization aspects like the number of sentences to be predicted and displayed. These configurations need to be in harmony with the prompt, as a single prompt alone is insufficient to fuel a comprehensive intelligent text entry system; the software function must align with these prompts. 
However, the in-depth details of the test keyboard are not covered in this paper due to the focus on other aspects.

\subsection{Final Output from \textit{Promptor}}

\begin{figure*}[t]
    \centering
    \includegraphics[width=0.8\textwidth]{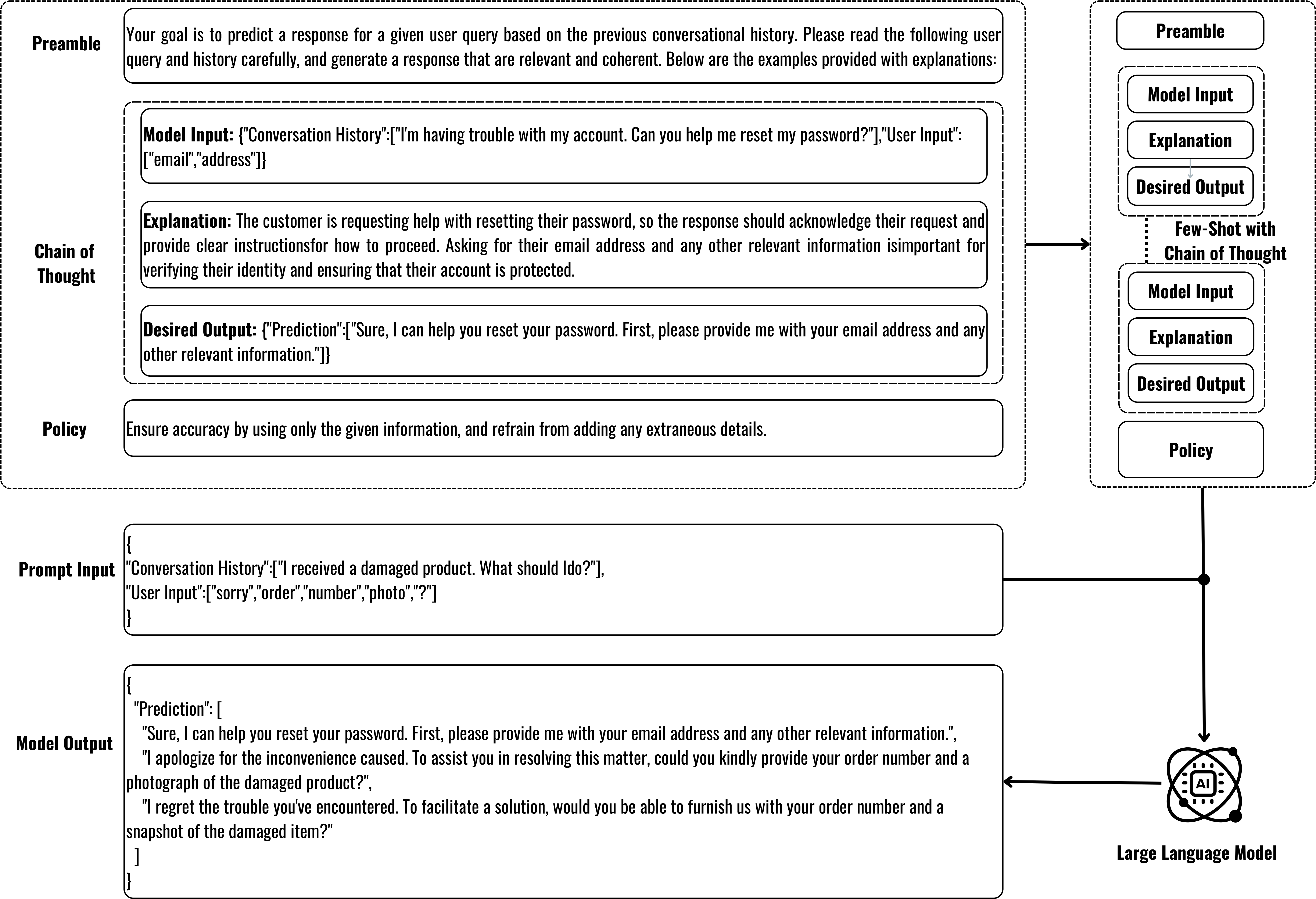}
    \caption{\textit{Promptor} ultimately will give the designer a final ``child'' prompt which is a \textit{system prompt} to guide the large language model to perform intelligent text entry tasks given an input. The \textit{system prompt} follows a structure consisting of Preamble, Few-Shot with Chain of Thought, and Policy.}
    \label{fig:prompt_output}
\end{figure*}  

Following the interactive session with the designer, \textit{Promptor} finalizes a ``child'' \texttt{system prompt} tailored to guide a GPT for the intelligent text entry function. An exemplar of this crafted prompt by \textit{Promptor} is depicted in Figure~\ref{fig:prompt_output}. Adhering to the guidelines established during the workshop, this \texttt{system prompt} encompasses instructions, examples, and policies. The instruction acts as an introductory preamble, setting a clear overarching objective. Examples are furnished using the few-shot approach, intertwined with chains of thought, showcasing sophisticated prompting techniques. From our preliminary experiments, we have selected to instruct \textit{Promptor} in the ``parent'' prompt to include four examples that strike a balance between optimizing output performance and not exceeding the context window length in the ``child'' \texttt{system prompt}. Additionally, it includes a policy part that delineates certain constraints to regulate the expansive capabilities of the language model. 

Each time the text prediction function of the large language model is invoked, the \texttt{user message} prompt to the model is presented in a JSON format. Similarly, the output from the model is also delivered in a JSON format, facilitating easy parsing to serve the text prediction function and integration into the text entry interface, as depicted in Figure~\ref{fig:teaser}. While the prompt template in the teaser figure serves as a basic illustrative example, in practice, the more intricate prompt template, as shown in Figure~\ref{fig:prompt_output}, would be employed. This underscores the significance of \textit{Promptor}'s ability to autonomously craft prompts. Designers unfamiliar with prompt engineering techniques would likely find the creation of such complex prompts daunting.

\section{Fine-tuning a Large Language Model or Prompting a Larger Language Model?}
The aim of this section is to underscore the limitations and suboptimal performance of fine-tuning large language models like GPT-2, especially when compared to simply prompting larger models such as GPT-3.5 or GPT-4.
Prior to the advent of GPT-3.5, a model renowned for its in-context learning capabilities, employing large language models for specific tasks necessitated fine-tuning. This fine-tuning process involved data collection, potential modifications to the model's input and output structures, and the use of high-performance computing resources. Such an approach demanded significant human effort, programming expertise, and computational power.
However, with the introduction of GPT-3.5, specific tasks can be addressed merely by prompting the model. This makes the deployment of large language models in specialized domains both faster and more efficient.
A pertinent question then arises: How does the performance of a Prompted GPT-3.5 stack up against a Fine-tuned GPT-2 for a specific task?

To shed light on this, we reference the study by Shen et al.~\cite{shen2022kwickchat}, where a GPT-2 model was fine-tuned for a text entry task for non-speaking users with motor disabilities. This task was designed to predict response sentences for these users during a conversation with a speaking partner, taking into account a set of keywords, the conversation history, and persona details. Impressively, this fine-tuned model surpassed traditional models, such as those rooted in information retrieval, in terms of sentence prediction accuracy.
In contrast, we will evaluate the performance of GPT-3.5 when prompted for the same task and compare the results.

To provide a more comprehensive answer, we have enhanced Shen et al.~\cite{shen2022kwickchat}'s system by implementing several pivotal modifications, each substantiated by improved experimental outcomes. Moreover, during our evaluation of both models, we adopted two distinct assessment methods: human evaluations, where human judges were employed, and AI evaluations, where we utilized GPT-4 as the judge.

\subsection{Evolving from KWickChat}
\label{sec:kwickchat}
We implemented certain modifications to enhance the performance of the KWickChat model to make it KWickChat~2. To support the reason for modification, we also report the performance improvement compared to the original KWickChat version in the evaluation metrics reported by Shen et al.~\cite{shen2022kwickchat}:
\begin{s_itemize}
    \item \textbf{Enabling Punctuation Input (Fine-Tuning GPT-2)}: 
    The original implementation of KWickChat described uses bag-of-keywords as the input to the model in order to generate high-quality responses. 
    However, text-based dialogues contain both words and punctuation to convey semantic meanings.
    The original KWickChat language generation model does generate responses that contain punctuation but does not allow the entry of punctuation by the user. 
    Reflecting on this limitation, we hypothesize that there may be a performance benefit in allowing users to explicitly input punctuation since punctuation can help efficiently convey semantic meaning, even if only combined with keywords.
    For example, an entry such as, ``favorite color ?'' can help prompt the system to generate the phrase, ``What is your favorite color?'' in preference to, ``My favorite color is blue.''

    We constrain the set of punctuation that can be inputted to avoid excessive cognitive burden on users, and allow users to complement the bag-of-keywords with four forms of punctuation: period, question mark, exclamation and comma.
    To support punctuation input for KWickChat, we first extracted the permitted set of punctuation from the golden reply.
    We then injected the extracted punctuation into the extracted keywords such that their order in the original sentence was preserved.
    This represents the user's new input to the language generation model and KWickChat was retrained on this basis. 

    We found that the general impact of including punctuation as input is placing additional constraints on the generated response, which in turn improves the quality of the responses. 
    We see a 79.3\% improvement in Bilingual Evaluation Understudy Score (BLEU)~\cite{papineni2002bleu} compared to when there is no punctuation used as the input (when the size of the bag-of-keywords is equal to 3).
    In practice, entering punctuation does require a keystroke but this is much less than the number of keystrokes required to input a word.  
    
    \item \textbf{Re-ranking Generated Responses (Post-Processing)}: 
    The extra language modeling head in the KWickChat language generation model is designed to constrain the syntax of the generated responses. 
    Despite this, some generated responses still contain obvious grammatical errors.
    
    To improve the quality of the sentence suggestions presented to users, we introduce a re-ranking mechanism that sorts the responses based on their score under a secondary language model. 
    More specifically, we use the original GPT-2 language model to perform scoring for the grammatical correctness of the sentence. 
    We use the exponential of the cross-entropy loss of GPT-2's language modeling head as the score.
    The GPT-2 language model we use as the scoring system is the original GPT-2 without any modification.
    It shares a similar structure to the KWickChat language generation model but has not been fine-tuned on the ConvAI challenge dataset. 
    Therefore, this secondary GPT-2-based scoring model serves as a totally distinct model from the KWickChat language generation model.
    
    This mechanism allocates scores to the \textit{M} generated responses and then sorts to them to obtain the top \textit{N} responses, which are then displayed as sentence suggestions.
    Note that \textit{M} is always larger than or equal to \textit{N}.
    The advantage of this re-ranking mechanism is most pronounced when \textit{M} is much larger than \textit{N}.

    We set \textit{M} to be 50 after considering the trade-off between the total prediction time of \textit{M} responses and the quality of the final re-ranked and filtered responses. 
    Our findings indicate that with three displayed sentence suggestions and using the re-ranking mechanism, the average keystroke savings increase by 15.9\% relative to the condition with no re-ranking. 
    Keystroke savings refers to the percentage of keystrokes that can be saved, as detailed in~\cite{shen2022kwickchat}. 
    We employ metrics distinct from BLEU as this re-ranking mechanism operates as a system-level add-on function beyond the language model; therefore, we utilize system-level measurements.
    As the number of sentence suggestions increases above eight, the improvement introduced by the re-ranking mechanism diminishes as eight alternatives without re-ranking are already likely to contain very high-quality responses.   
\end{s_itemize}


\subsection{Prompting GPT-3.5}
Our goal is to prompt GPT-3.5 to equip it with the core feature of predicting multiple sentence candidates based on conversation history, persona, and a user's set of keywords. Additionally, it should have the advanced capability of incorporating punctuation for users. In devising the optimal approach, we consulted with \textit{Promptor} and finalized a specific prompt. We utilized this prompt to configure GPT-3.5 as KWickChat 2. Remarkably, this entire procedure was completed in under 20 minutes, showcasing its efficiency, especially when compared to the time-intensive process of fine-tuning a KWickChat model for system integration. Implementing GPT-3.5 is streamlined, necessitating only API calls.

\subsection{Fine-tuning GPT-3.5}
OpenAI has recently introduced the fine-tuning functionality for GPT-3.5. 
We employed the OpenAI fine-tuning APIs to fine-tune GPT-3.5, utilizing the same prompt as we did for the Prompted GPT-3.5 version. The fine-tuning process follows the same training and testing split from Shen et al.\cite{shen2022kwickchat}.

\subsection{Joint Analysis from Human and AI Judgments}
To assess the performance of both models, we utilize two evaluation methods: Human Judgment and AI Judgment. 
A wealth of research has delved into the potential of AI for streamlining the evaluation of NLP tasks~\cite{kocmi2023large,chiang2023can}. Evaluations conducted by humans interpret model outputs with a depth of understanding, providing rich, qualitative feedback that captures the intricacies of user interactions. This method, though versatile, carries the potential for subjectivity and might face scalability challenges. Conversely, AI-driven evaluations benchmark model performance against predefined standards, ensuring results that are both objective and consistent. Moreover, such evaluations offer the advantages of scalability and cost-effectiveness.
We use this dual-method approach to serve dual purposes. Firstly, it allows us to evaluate the models from two distinct perspectives, ensuring a more balanced comparison. Secondly, it enables us to investigate the correlation between human judgment scores and AI judgment scores. Understanding this relationship can pave the way for primarily relying on AI judgment in large-scale evaluations for future assessments.

\subsubsection{Human Judgements}
Following the human judgment analysis strategy of Shen et al.~\cite{shen2022kwickchat}, we enlisted two native English-speaking volunteers as judges. These judges were tasked with evaluating the quality of the generated responses, using a scale from 1 (very bad) to 5 (very good), with the golden reply serving as a benchmark. The conversation excerpts provided to the judges also incorporated persona details for context.

The evaluation process was structured as follows:
The initial conversation served to elucidate the evaluation procedure.
The subsequent five conversations were earmarked for practice.
The core evaluation centered on the remaining 25 conversations.
Each of these conversations comprised four exchanges. An exchange consisted of an utterance from the speaking partner, a reference reply, and four generated responses presented as suggested candidates. The two judges independently carried out their evaluations. Subsequently, their scores for each prediction were averaged. To ensure fairness, the human judges evaluated responses from the two models in a counterbalanced sequence.

\subsubsection{AI Judgements}
For AI-based evaluations, we employed OpenAI Evals, a framework designed for assessing LLMs (large language models) and systems that incorporate LLMs as integral components~\cite{OpenAI2023}. We Prompted GPT-4 with an introductory preamble outlining the evaluation criteria based on the provided information, mirroring the instructions given to human evaluators for the initial five examples. GPT-4 then proceeded to score the subsequent 25 conversations. This process was repeated five times using different random seeds, and the scores were subsequently averaged.
It's important to note that we employ GPT-4 as the judge, while GPT-3.5 is used for performing the intelligent text prediction. This approach helps to minimize potential bias, thereby preventing a situation akin to "the judge being the sportsman at the same time". Furthermore, the prompts used to guide GPT-4 and GPT-3.5 are distinctly different, which further aids in minimizing any bias.

\subsubsection{Results}


\begin{table*}[t]
\centering
\caption{Comparative analysis of the mean and standard deviation (indicated as Mean(±Std Dev)) of judgments made by humans and an AI across three different models: Fine-tuned GPT-2 (with post-processing), Prompted GPT-3.5, and Fine-tuned GPT-3.5. The table also presents the percentage improvement in human and AI judgments from Fine-tuned GPT-2 to Prompted GPT-3.5, and from Prompted GPT-3.5 to Fine-tuned GPT-3.5, along with the p-value of the Wilcoxon Statistic indicating the statistical significance of these improvements.}
\label{tab:my_label}
\begin{tabular}{lcp{2cm}p{2cm}cp{2cm}}
\toprule
\textbf{Metric} & \textbf{\begin{tabular}[c]{@{}c@{}}Fine-tuned \\ GPT-2\end{tabular}} & \textbf{\begin{tabular}[c]{@{}c@{}}Prompted \\GPT-3.5\end{tabular}}& \textbf{\begin{tabular}[c]{@{}c@{}}Fine-tuned \\ GPT-2 \\to \\Prompted \\GPT-3.5\end{tabular}} & \textbf{\begin{tabular}[c]{@{}c@{}}Fine-tuned \\GPT-3.5\end{tabular}} & \textbf{\begin{tabular}[c]{@{}c@{}}Prompted \\GPT-3.5 \\ to \\Fine-tuned \\GPT-3.5\end{tabular}} \\
\midrule
\begin{tabular}[c]{@{}c@{}}Human Judge\\Mean (±Std Dev)\end{tabular} & 4.27(±0.82) & 4.31(±0.79) & 0.99\% (p=0.0) & 4.29(±0.80) & -0.57\% (p=0.11) \\
\begin{tabular}[c]{@{}c@{}}AI Judge\\Mean (±Std Dev)\end{tabular}   & 3.96(±0.99) & 4.23(±0.86) & 6.86\% (p=0.0) & 4.26(±0.85) & 0.71\% (p=0.09) \\
\bottomrule
\end{tabular}
\label{tab:human_ai}
\end{table*}

\begin{figure}[!ht]
    \centering
    \subfloat[\centering Fine-tuned GPT-2 (with post-processing)]{{\includegraphics[width=4.5cm]{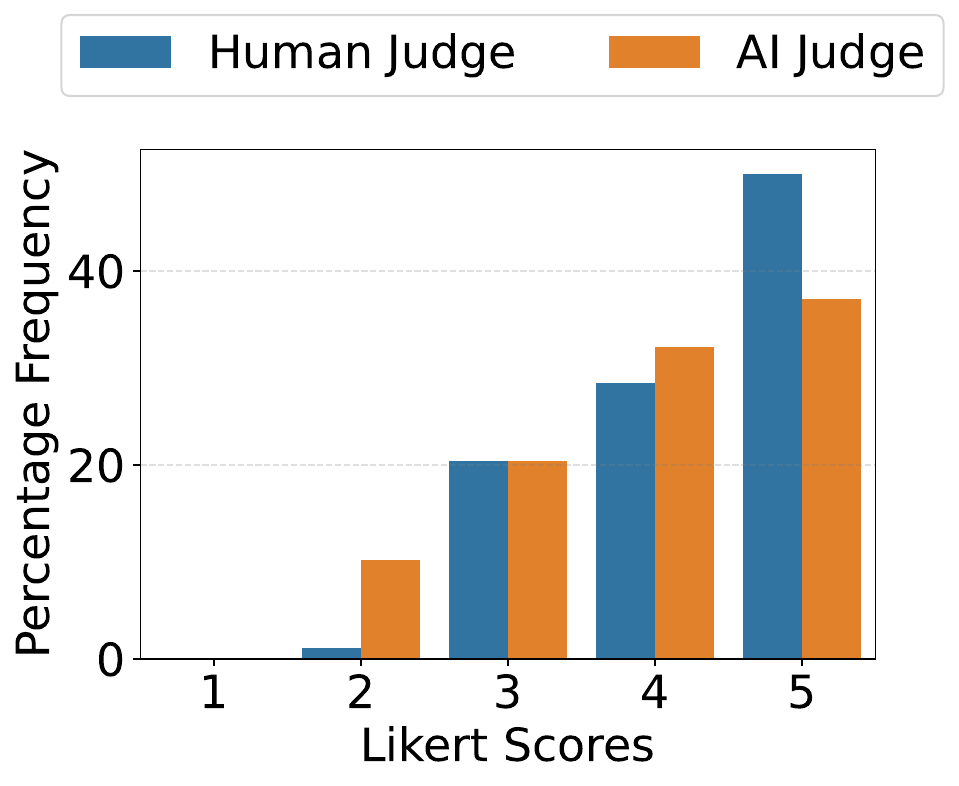} }}%
    \qquad
    \subfloat[\centering Prompted GPT-3.5]{{\includegraphics[width=4.5cm]{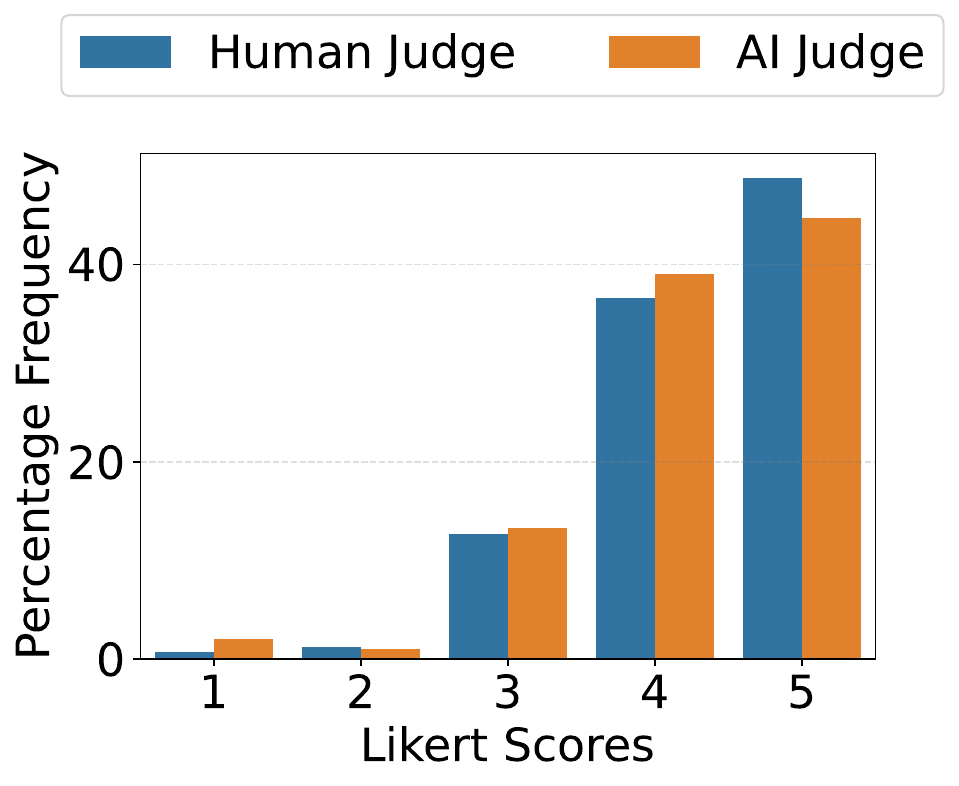} }}%
        \qquad
    \subfloat[\centering Fine-tuned GPT-3.5]{{\includegraphics[width=4.5cm]{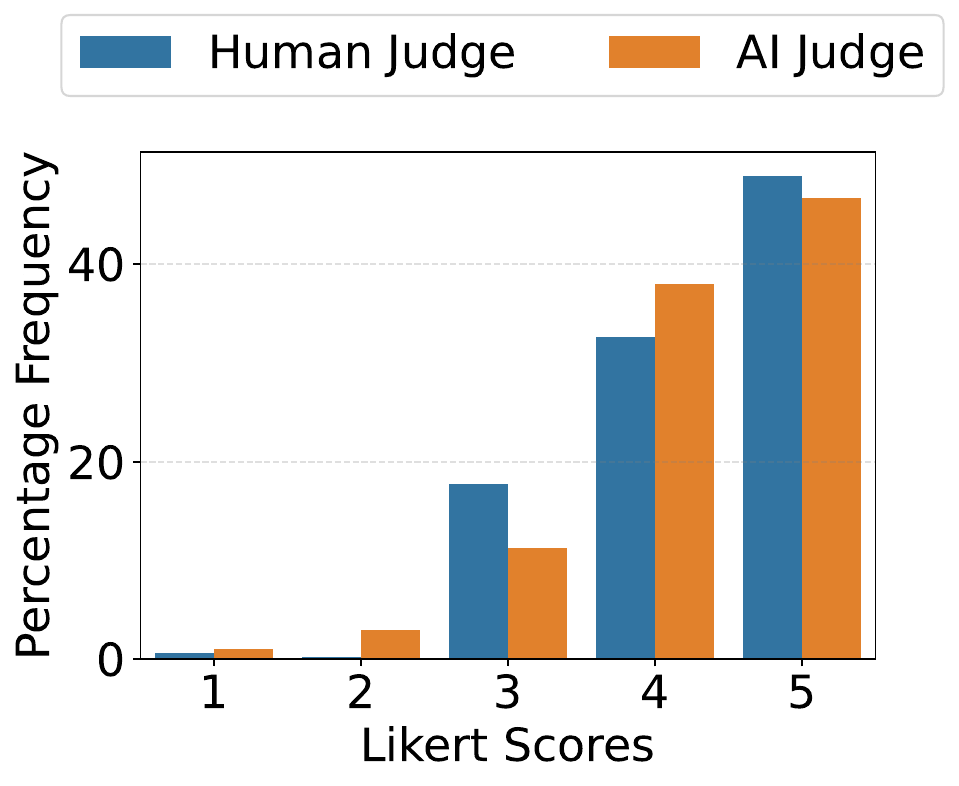} }}%
    \caption{Bar chart illustrating the comparative scores from human and AI evaluations for the Fine-tuned GPT-2 model (with post-processing), the Prompted GPT-3.5 model, and the Fine-tuned GPT-3.5 model.}%
    \label{fig:human_ai}%
\end{figure}

\begin{table}[!ht]
\centering
\caption{Spearman's Correlation and Cohen's Kappa of the scores given by the human judge and AI judge for different models and for all models combined together.}
\label{tab:my_label}
\begin{tabular}{lcc}
\hline
\textbf{Scores}               & \textbf{\begin{tabular}[c]{@{}c@{}}Spearman's\\Correlation\end{tabular}} & \textbf{\begin{tabular}[c]{@{}c@{}}Cohen's\\Kappa\end{tabular}} \\ \hline \\
Fine-tuned GPT-2    & 0.88                   & 0.54          \\
Prompted GPT-3.5    & 0.94                   & 0.87          \\
Fine-tuned GPT-3.5  & 0.96                   & 0.86          \\
Combined & 0.91                   & 0.75          \\ \hline
\end{tabular}
\label{tab:correlation}
\end{table}

\begin{s_itemize}
    \item \textbf{The Inferior Performance of Fine-tuned GPT-2 Does Not Justify the Associated Data Collection and Expensive Model Development.}

Table~\ref{tab:human_ai} indicates the average scores and their standard deviations rated by Human Judge and AI Judge. We see obvious higher scores of Fine-tuned GPT-2 compared with Prompted GPT-3.5. 
The results of Wilcoxon Signed-Rank Test~\cite{woolson2007wilcoxon} provide statistically significant evidence that Prompted GPT-3.5 outperforms Fine-tuned GPT-2 in the task of intelligent text input. This conclusion is supported by both human and AI evaluations. Both the human and AI evaluations yielded a p-value of zero, indicating a significant difference. Therefore, we can confidently assert that the difference between GPT-2 and GPT-3.5 in intelligent text entry is not only statistically significant but also likely to be practically meaningful. This underlines the effectiveness of prompting GPT-3.5 for tasks involving intelligent text input.

Figure~\ref{fig:human_ai} showcases bar plots that depict the distribution of scores from both human and AI judges. 
We noticed that human judges tend to assign higher scores, particularly for the Fine-tuned GPT-2. In a follow-up interview with the human judges, they explained that their awareness of the system's intended use for people with disabilities led to greater tolerance in their evaluations. For example, GPT-2 sometimes produces sentences that, while conveying the intended meaning, may contain punctuation or structural errors. Human judges were more forgiving of these issues due to the fact that this system is used for disabled people, and they are already impressed by the promising results already achieved by a sentence prediction system. 
On the other hand, GPT-4, serving as the AI judge, was also provided with the same background knowledge. However, its scores were more objective, reflecting the inherent advantage of AI in maintaining impartiality.

 \item \textbf{Zero Improvement of Fine-Tuning GPT-3.5 Does Not Justify the Associated Data Collection and Expensive Model Request Calls.}

Table~\ref{tab:human_ai} and Figure~\ref{fig:human_ai} indicate that the Fine-Tuned version does not outperform the Prompted GPT-3.5. Therefore, the result does not justify the time invested in data collection, especially in scenarios with limited training data. Although the fine-tuning process via OpenAI APIs is relatively straightforward, requiring only a few lines of code, the most challenging aspect lies in the data collection and cleaning to prepare the data in a format suitable for fine-tuning through the OpenAI APIs.
Additionally, fine-tuning GPT-3.5 comes at a financial cost, with the request for a Fine-tuned GPT-3.5 model being approximately eight times more expensive than employing the basic models. 

Moreover, being able to fine-tune GPT-3.5 doesn't rule out prompt engineering for specific needs. As per OpenAI's guidance, starting with prompt engineering is advisable for better initial results. This approach, quicker than the fine-tuning process, still holds value even if fine-tuning becomes necessary later. It often yields the best results when a well-crafted prompt is used in fine-tuning. Nonetheless, designers can utilize \textit{Promptor} to identify potent prompts, and if necessary, fine-tune the model using their dataset, should one be available.

\item \textbf{Human Judge and AI Judge Have a High Level of Alignment and Agreement.}

We performed Spearman's correlation~\cite{hauke2011comparison}  for the score list of different models for the two groups, human judges and AI judges, and also calculated the stats for the combined score lists. Table~\ref{tab:correlation} shows that the Spearman's correlation values range from 0.88 for Fine-tuned GPT-2, 0.94 for Prompted GPT-3.5, to 0.96 for Fine-tuned GPT-3.5, and 0.91 for the combined score list, indicating a strong monotonic relationship between human and AI judgments across different models. 

When considering the human and AI judges as two independent raters, we can apply Cohen's Kappa~\cite{kvaalseth1989note}, a measure of inter-rater reliability that accounts for the possibility of random agreement. In Table~\ref{tab:correlation}, a Cohen's Kappa value of 0.54 for Fine-tuned GPT-2 suggests moderate agreement. This was anticipated as human judges tend to give ``sympathetic'' scores on grammar and structure errors in predictions, considering the use case for disabled users, while AI provides objective scores. On the other hand, for the other two models, we observe that the Cohen's Kappa values, ranging from 0.86 to 0.87, demonstrate a strong level of concordance between the human and AI judges. Overall, a Cohen's Kappa score of 0.75 suggests substantial agreement, reflecting a high degree of alignment in judgments between the human and AI judges across models.

Given this strong alignment and agreement, AI judgments can be reliably employed for large-scale evaluations in similar tasks. In subsequent sections, we utilize AI judgment to assess the performance of prompts in large scale, whether crafted by designers or generated by \textit{Promptor}.

\end{s_itemize}

\section{Evaluation of \textit{Promptor} with Designers}

\begin{table*}[t]
\centering
\caption{Conversation examples for the three use cases. Agent input is extracted from dialog act to represent the text input from the agents. }
\begin{tabular}{p{6cm} p{6cm} p{2cm}}
\toprule
\textbf{Conversation} & \textbf{Dialog Act} & \textbf{Agent Input} \\ 
\midrule
\hline
\multicolumn{3}{c}{Movie-Ticket Booking Task} \\ \hline
\midrule
usr: Find me a good action movie this weekend. &  &  \\ 
\midrule
agt: London Has Fallen is currently the number 1 action movie in America. & agt: inform(moviename=london has fallen; other=number 1; genre=action) & london has fallen number 1 action \\ 
\midrule
usr: Oh that sounds terrific &  &  \\ 
\midrule
agt: Would you like to purchase tickets to this movie? I would need to know what city you are in. & agt: request(city) & city ?\\ 
\midrule
\hline
\multicolumn{3}{c}{Restaurant Reservation} \\ \hline
\midrule
usr: I'm looking for a martini bar in Indianapolis. &  & \\ 
\midrule
agt: Here is the restaurant I found: High Velocity. Do you want to book? & agt: request(reservation;restaurantname=High Velocity) & High Velocity? \\ 
\midrule
usr: YES &  & \\ 
\midrule
agt: at what date would you like to go? & agt: request(date) & date ?\\ 
\midrule
\hline
\multicolumn{3}{c}{Taxi Booking} \\ \hline
\midrule
usr: I would like to book a cab please. & &  \\ 
\midrule
agt: On what date would you like a taxi? & agt: request(date) & date ? \\ 
\midrule
usr: today &  &  \\ 
\midrule
agt: How many are going? & agt: request(numberofpeople) & number of people ?\\ 
\bottomrule
\end{tabular}
\label{tab:example_conv}
\end{table*}

Our findings in Section~\ref{sec:kwickchat} confirm that prompting GPT-3.5 to harness intelligent text entry functionality is not only efficient but also outperforms the Fine-tuned GPT-2. This underscores the significance of utilizing GPT-3.5 prompts for intelligent text input.
Consequently, we need to rigorously evaluate our proposed \textit{Promptor}, particularly in comparison with designers lacking extensive prompt engineering experience. We conducted a user study involving two groups of text prediction system and conversational agents designers, none of whom have a strong background in prompt engineering. These designers were tasked with prompting GPT-3.5 to acquire text prediction functionality across three use cases, each with publicly available datasets for ease of evaluation. Group A self-crafted prompts, while Group B used \textit{Promptor} to generate prompts.

\subsection{Experiments}
Overall, our study was guided by two objectives. Firstly, we compare the efficacy of prompts produced by \textit{Promptor} and the prompts created by the designers. We employed OpenAI Eval~\cite{OpenAI2023}, to assess these prompts' effectiveness across three distinct use cases using an open dataset.
Secondly, we assess the user experience and usability of \textit{Promptor}. We sought to determine if it offers a positive interaction for designers and if its generated prompts significantly outperform those crafted by the designers themselves. To measure this, we utilized the System Usability Scale (SUS) and gathered post-study feedback.

\begin{s_itemize}
    \item \textbf{Example Use Case}:
    Our study employs three distinct use cases, each featuring public conversational datasets with unique characteristics: movie-ticket booking, restaurant reservation, and taxi booking~\cite{li2018microsoft}. Examples from these datasets are presented in Table~\ref{tab:example_conv}. A key advantage of these datasets is the inclusion of Dialogue Actions, which represent actions from both the user and the agent. These actions, essentially human annotations, facilitate the training of conversational agents. 
    While most conversational datasets were originally devised to advance the development of conversational robots (chatbots), the majority of public conversational datasets lack inputs from dialogue partners. This omission poses challenges in developing and evaluating conditional conversational robots, which essentially function as intelligent text entry systems. 
    Given GPT-3.5's ability to process unstructured data and natural language (i.e., data not in dialogue actions), we convert the dialogue actions back into natural language. By evaluating prompts across these use cases, we can gain a comprehensive understanding of the potential benefits and constraints of the prompts in various scenarios.
    \item \textbf{Participants}: We recruited 24 designers well-experienced in sentence prediction systems, with an average age of 29.3 and a standard deviation of 4.2. The group comprised 17 males and 7 females. Among them, six designers had over 5 years of experience with sentence prediction systems, such as customer service chatbots, while ten had more than 3 years of expertise. The remaining designers had less than 3 years of experience in the field. Notably, none of them had substantial experience in prompting large language models or in using such models to build applications. The number of designers was determined using the G*Power analysis tool~\cite{faul2007gpower}. 

    \item \textbf{Procedure}: 
    Participants were initially introduced to the task of crafting prompts to guide GPT-3.5 in performing intelligent sentence prediction functions. The model output should be in JSON format and contain four predictions. The overarching goal was to enhance the accuracy of prediction responses by leveraging all available information, whether it pertained to personalization, predictions based solely on input, or drawing from previous histories. Therefore, it is the participants's choice to select which available information to use for text prediction. 
    To provide context, participants were shown three use cases along with sample datasets related to each.
    The 24 participants were then randomly divided into two groups.

    Group A relied on their intuition to design a prompt generator for each case. While they didn't have access to \textit{Promptor}, they were free to search the internet for guidance.
    Conversely, Group B was acquainted with \textit{Promptor} and guided on how to employ it for prompt creation.
    In both groups, participants were encouraged to vocalize their thought processes and articulate their ideas, either through text or rudimentary sketches.
    All participants crafted prompts for the three use cases, with the sequence of use cases randomized for each individual.
    A time frame of 30 minutes was allocated for each participant to iteratively develop their prompt.
    During the session, participants could test their crafted prompts using a test keyboard, as depicted in Figure~\ref{fig:test_kb}.

    To evaluate the effectiveness of \textit{Promptor} for Group B, we utilized the System Usability Scale (SUS) and complemented it with post-study questionnaires. These questionnaires were designed to gather quantitative feedback on participants' experiences with \textit{Promptor}, asking questions such as "What aspects of \textit{Promptor} do you appreciate the most?" and "What improvements would you suggest for \textit{Promptor}?"
    For Group A, we provided a post-study questionnaire that focused on their experiences with writing prompts. The questions included ``What are your thoughts on the process of writing prompts?''.

    \item \textbf{Materials}: We use a Windows machine to operate the interactive GUI for \textit{Promptor} and to host the virtual keyboard that features the text prediction testing functionality.
\end{s_itemize}

\subsection{Evaluation}

\subsubsection{Evaluation Metrics}
Contrary to the overall quality score metric used in Section~\ref{sec:kwickchat}, our aim here is to delve deeper into the nuances of model predictions.
Firstly, we introduce a new metric, \textit{Format Correctness}. Without proper prompting and regulation, the model may yield unstable outputs, not always adhering to the JSON format. Other formats, such as "your predicted responses is: ...", may appear. Some model outputs may not even be the desired predictions but may instead present clarifications, like "Could you please clarify your request?" All these instances violate format correctness. We initially measure format correctness using a rule-based system, considering any output that cannot be parsed by a Python script for JSON format as format incorrect. 
We first use this metric to filter out responses with incorrect formatting.

Subsequently, we employ two additional evaluation metrics that specifically assess the quality of the predictions: similarity and coherence.
The \textbf{Similarity} metric evaluates whether the predicted response encapsulates the main information and context from the original response. A higher similarity score is indicative of high-quality responses. However, some low similarity scores may not necessarily reflect poor model performance, but rather, they could be due to the user input in the dataset not conveying all the necessary information.
The reason we opt for GPT-4 to score \textbf{Similarity} over BLEU is due to GPT-4's potential to discern more nuanced and semantic similarities between sentences, courtesy of its advanced deep learning architecture. On the other hand, BLEU operates at a more superficial level, concentrating on exact string matches and n-gram overlaps, which might not accurately reflect the semantic similarity between text passages~\cite{reiter2018structured}.
Therefore, relying solely on similarity may not fully capture the overall quality of the predictions. To address this, we introduce the \textbf{Coherence} metric. It assesses whether the predicted sentence is coherent with the user's request, the agent's dialogue actions, and the booking process. By employing these two distinct metrics, we can separately scrutinize the quality of the predicted response from different perspectives.

\subsubsection{Large-Scale AI Evaluation}
As identified in Section~\ref{sec:kwickchat}, there is a strong agreement between AI and human judges. Furthermore, the large-scale evaluation of dozens of prompts for GPT-3.5 across multiple extensive datasets makes it impractical to solely rely on human judges for evaluations. Consequently, we utilize OpenAI Evals and Prompted GPT-4 to measure the similarity and coherence of the responses using a Likert scale. 
Employing AI judges introduces more objective and consistent outcomes. Since our focus is on the relative comparison between two groups rather than the absolute quality of the responses, the advantages of AI evaluation - scalability, objectivity, consistency, and cost-efficiency - can be fully leveraged in this context.

\subsection{Results}

\begin{figure*}[t]
  \centering
  \begin{minipage}[b]{0.5\linewidth}
    \includegraphics[width=\linewidth]{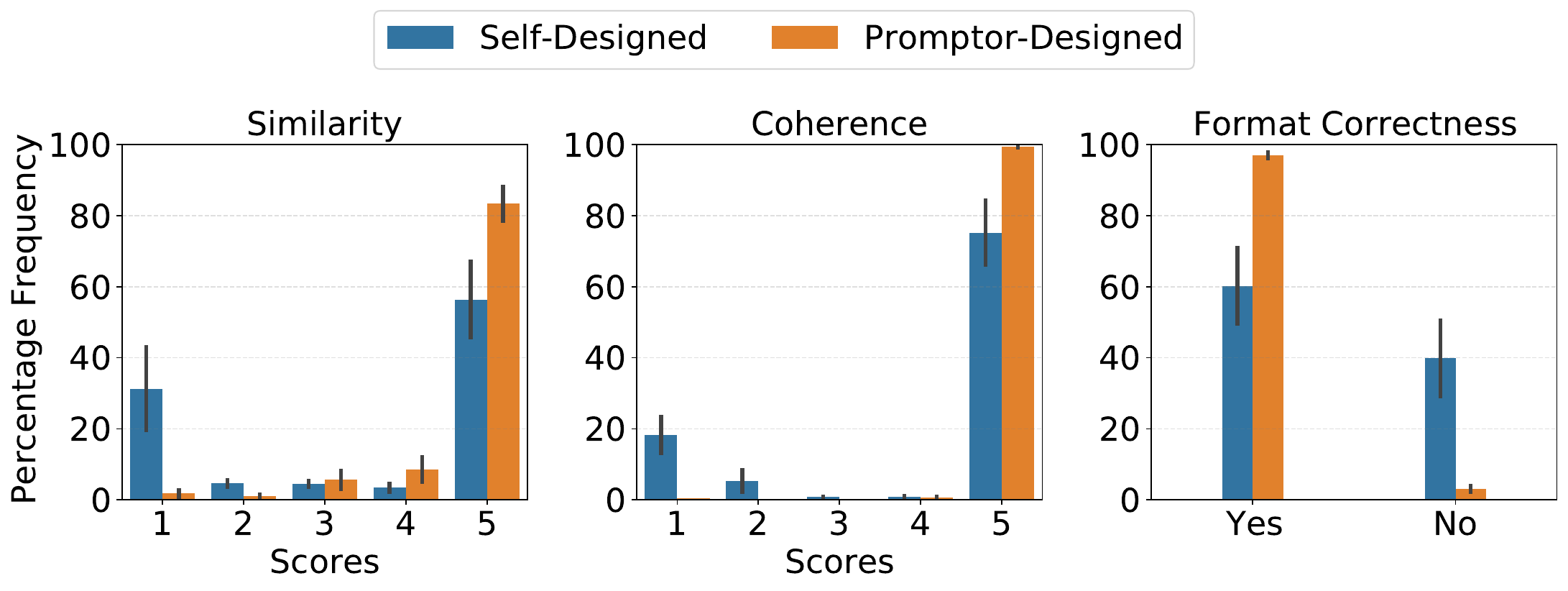}
    \subcaption{Taxi Booking}
    \label{fig:study2_technical}
  \end{minipage}
  \hfill
  \begin{minipage}[b]{0.5\linewidth}
    \includegraphics[width=\linewidth]{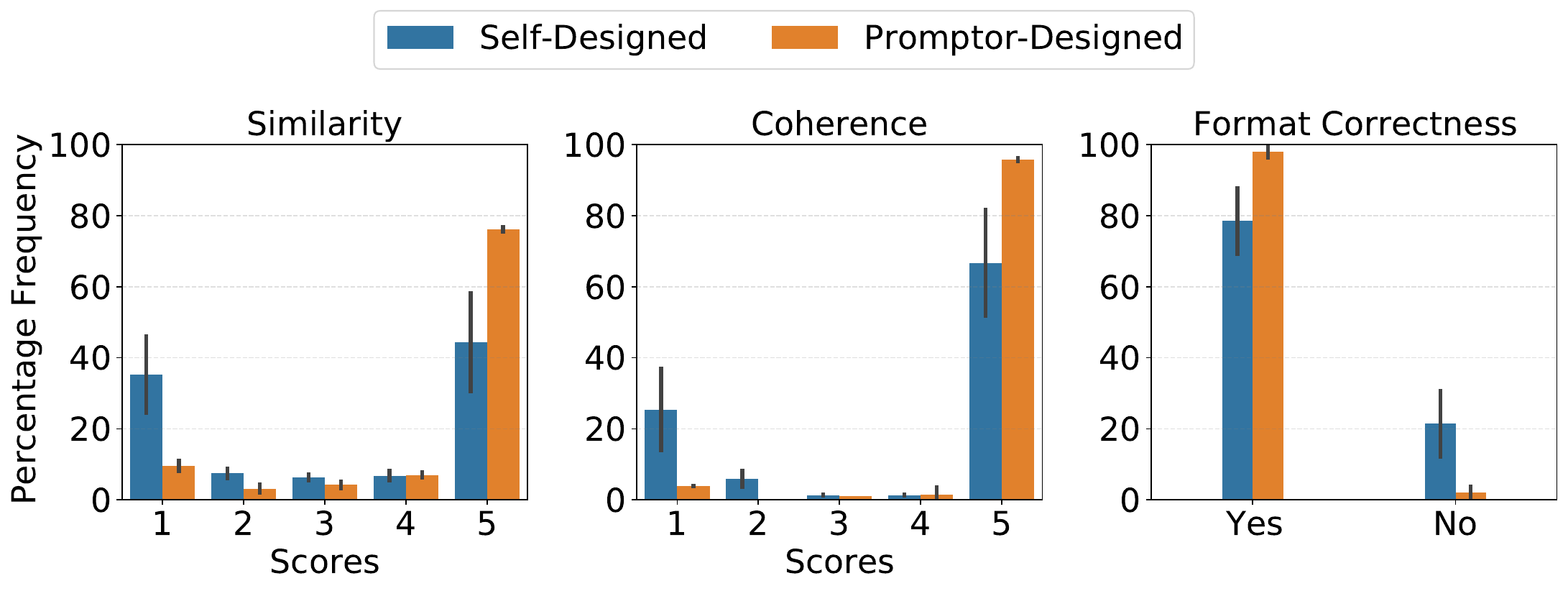}
    \subcaption{Movie-Ticket Booking}
    \label{fig:study2_movie}
  \end{minipage}
  \hfill
  \begin{minipage}[b]{0.5\linewidth}
    \includegraphics[width=\linewidth]{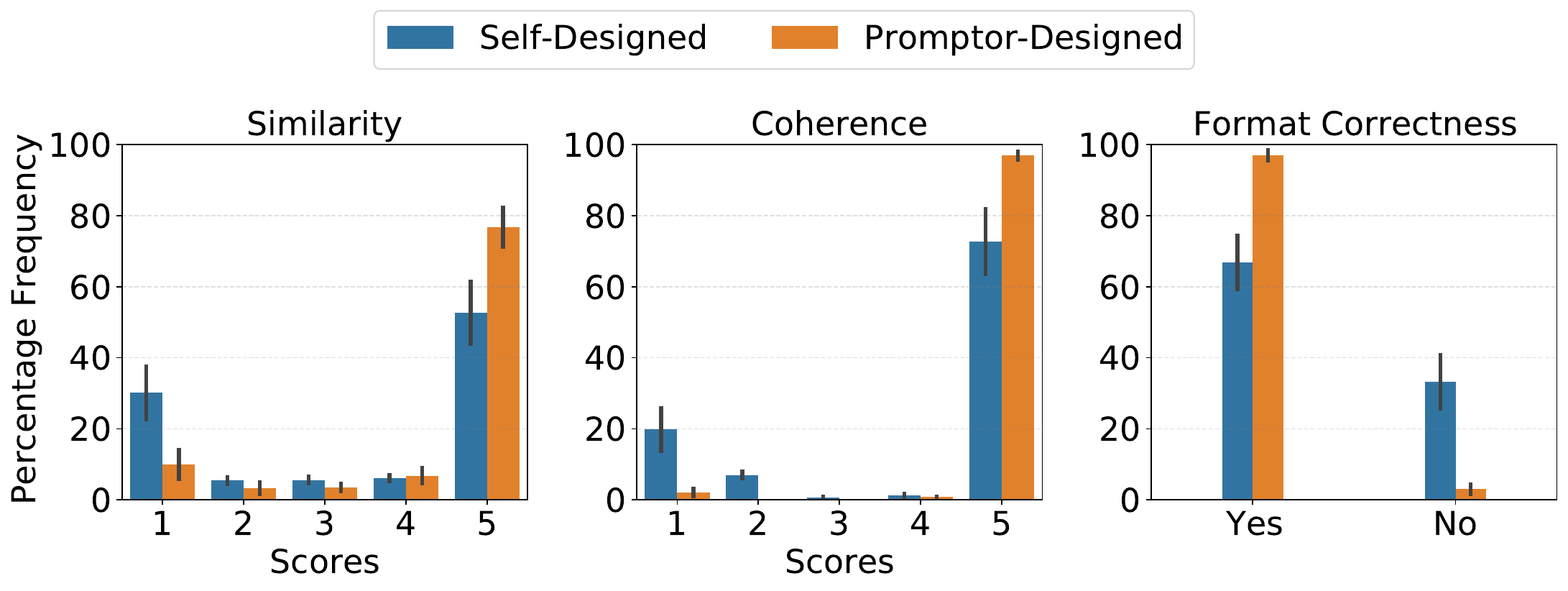}
    \subcaption{Restaurant Reservation}
    \label{fig:study2_rest}
  \end{minipage}
  \caption{Comparison of prompt generators from Group A (Self-Designed) and Group B (Promptor-Designed). The evaluation encompasses three metrics: similarity, coherence, and format correctness. While the first two metrics are assessed using a Likert scale, the format correctness is determined with a binary `yes' or `no' response. The outcomes highlight the superior performance of the Promptor-Designed prompts, underscoring the advantages of employing \textit{Promptor}.}
    \label{fig:prompt_comparison}
\end{figure*}

\subsubsection{\textit{Promptor} received positive feedback from designers.}
The System Usability Scale (SUS) score averaged at 83.4, earning an "excellent" rating. This score reflects the participants' appreciation for the system.


Feedback from the post-study questionnaire highlighted participants' appreciation for \textit{Promptor}. \textit{P2} noted, ``Promptor's generated prompts perfectly fulfill my needs. \textit{Promptor} quickly adapts itself to my needs.'' \textit{P6} complimented \textit{Promptor} for its ``structured, modularized, and accurate prompts''. \textit{P4} praised \textit{Promptor} as an educational tool, saying, ``I learned a great deal from \textit{Promptor}, like understanding different parameters and the distinction between system and user messages.'' \textit{P12} underscored \textit{Promptor}'s speed and efficiency in generating effective prompts, calling it a ``real time-saver.'' Lastly, \textit{P8} emphasized the user-friendliness of \textit{Promptor}'s web-based user interface, which mirrors the ChatGPT interface and offers options to select the model and adjust parameters.

Using the feedback from the participants, we categorized their positive responses about \textit{Promptor} into three main themes. These themes are associated with keywords that describe the system's attributes: 
\begin{s_itemize}
    \item \textbf{Effectiveness, Structured and Stable:} \textit{Promptor} excels in generating effective and well-structured prompts to consistently produce accurate predictions. (\textit{P1, P2, P3, P4, P5, P6, P7, P8, P9, P10, P11, P12})
    \item \textbf{Robustness, Creativity, Flexibility:} \textit{Promptor} was praised for its robust and flexible prompt generation capabilities, and its creative approach to addressing user queries. (\textit{P1, P2, P4, P6, P8, P10, P12})
    \item \textbf{Knowledgeability, Domain Expertise and Educational:} Participants appreciated \textit{Promptor}'s depth of knowledge and its domain expertise, which is reflected in its responses. (\textit{P2, P3, P4, P6, P7, P9, P11, P12}) 
    \item \textbf{Ease of Use, Simplicity and Straightforwardness, Time-Efficient:} The intuitive, simple, and straightforward conversational interface of \textit{Promptor} was praised by the users. (\textit{P1, P2, P5, P6, P7, P8, P9, P12}) 
\end{s_itemize}

Participants provided valuable feedback and suggestions for improving \textit{Promptor}. \textit{P2} proposed enhancing user experience by enabling voice interactions with \textit{Promptor}, stating, ``Such a conversational interface could be more accessible through speech.'' \textit{P8} suggested automating \textit{Promptor} further, saying, ``\textit{Promptor} should automatically implement the prompt and intelligent text prediction functions for me.''
\textit{P11} saw potential in \textit{Promptor} as a tool for both designers and end users, expressing, ``\textit{Promptor} should be made available to normal users for personalizing their text entry methods. As a designer, I'm open to learning proficient prompt engineering. However, for ordinary users, \textit{Promptor} could be an extremely useful tool.''
This feedback underscores \textit{Promptor}'s potential to democratize access to personalized text prediction for both designers and ordinary users. Particularly noteworthy were suggestions to automate the implementation and deployment of sentence prediction functions, which could empower users to personalize their own text prediction system.

\subsubsection{Designers without access to \textit{Promptor} find it challenging to construct good prompts.}

Crafting effective prompts for large language models can be a daunting task, especially for designers lacking expertise in prompt engineering. The task demands a comprehensive understanding of the model's workings and an ability to predict its responses to different prompts. 
\textit{S2} (Participant 2 from Group A) highlighted these challenges, stating, ``Creating a prompt independently was difficult. I often faced uncertainty on how best to communicate my intentions to ChatGPT. This experiment highlighted the complexities of crafting task-specific prompts. Formulating precise prompts that effectively guide GPT models is challenging.''
Similarly, \textit{S7} shared, ``I was uncertain about structuring the prompt due to the lack of a standard format. The hardest part was dealing with the unpredictability of GPT's responses and the uncertainty around optimizing the prompts. The same prompt could lead to varied results, causing unstable outputs.''
This designer further noted, ``I frequently had to experiment with multiple prompts before finding one that yielded satisfactory results.''

We similarly categorized the participants' feedback into three main themes.
 These included:
\begin{s_itemize}
    \item \textbf{Lack of Expertise:} With limited knowledge in prompt engineering, initiating the process seemed challenging. The absence of a clear starting point added to the complexity. (\textit{S1, S2, S3, S4, S5, S7, S9, S11, S12})
    \item \textbf{Stability and Consistency:} The designers found it challenging to guide the large language model to generate stable and consistent results that aligned with their prompts. (\textit{S1, S4, S6, S9, S11, S12})
    \item \textbf{Integration Difficulties:} Even when the designers managed to get the language model to generate predicted sentences, integrating these into the actual text entry system was a hurdle. Without a clear understanding of how to use placeholders in prompts, the practical implementation of the prompts became complex. (\textit{S3, S5, S6, S10, S11})
\end{s_itemize}

Despite these challenges, some designers found the process rewarding. For instance, \textit{s10} reported, ``Experimenting with different prompts can be intriguing. I feel confident that I could create effective prompts if I delve deeper into the process.'' From this, we can infer that while \textit{Promptor} may not be suitable for everyone, it can prove highly beneficial for those seeking to save time or for those who prefer not to delve into structured learning of prompt engineering.

\subsubsection{\textit{Promptor}-designed prompts generate better predictions than self-designed prompts. }
Figure \ref{fig:prompt_comparison} reveals a significant improvement in all three evaluation metrics when using \textit{Promptor}-designed prompts compared to self-designed ones. This improvement is especially notable in terms of format correctness. Self-designed prompts resulted in at least 30\% of predictions that did not meet the required format.

Without specific prompting, large language models might generate results in incorrect formats, use varied quotation marks, or even add unnecessary details such as ``your predicted sentence is:''. In some instances, they might not produce complete sentences but rather phrases, words, or even dictionaries that include the agent's input. This greatly hampers the performance of text prediction functions, as incorrect formatting cannot be parsed, rendering the sentence prediction non-functional on the keyboard.
In contrast, \textit{Promptor}-designed prompts generate stable outputs with the correct format, with an error rate of less than 5\%. The format correctness can be attributed to several factors, such as the use of correct syntax, examples, capitalization for emphasis, and constraint policies. Many designers were not aware of these prompt engineering skills, resulting in unstructured and informal natural language prompts.

Additionally, self-designed prompts attained an average similarity score of 2.91 and a coherence score of 3.41, whereas those designed by \textit{Promptor} garnered a superior average with a similarity score of 3.93 and a coherence score of 4.16, marking an enhancement of 35\% in similarity and 22\% in coherence respectively.
This suggests that when the user's input conveys most of the necessary information, the prompts can guide GPT-3.5 to produce highly accurate responses. Even when the user's input does not cover all necessary information, the predicted responses remain coherent with the background. In contrast, self-designed prompts do not guide GPT-3.5 adequately to produce coherent prediction responses, let alone similar responses. This can be attributed to several reasons: i) The designers did not adequately describe the scenarios, such as the user's goal and profile, and the data profile. As a result, GPT-3.5 could not comprehend the specific task, leading to incoherent predictions. ii) The designers did not leverage all the given information to prompt GPT-3.5 to make the predictions. Some users only used the user's input and neglected or ignored the previous conversation history. In contrast, \textit{Promptor} provides users with a comprehensive background of different perspectives of information to leverage. Notably, all users in Group B utilized both the conversational history and user's input.

\section{Discussion}
In this section, we primarily discuss the generalizability of \textit{Promptor}, including the scenarios where \textit{Promptor} is suitable, and its potential applicability in other further scenarios.

\subsection{Eliminating Data Collection and Model Development Needs with \textit{Promptor}}
We delve into two approaches for intelligent text predictions: Fine-tuned GPT-2 and Prompted GPT-3.5, each catering to different scenarios. The core advantage of Fine-tuned GPT-2 is its capability to operate locally. Thus, scenarios demanding local development, whether due to privacy considerations or internet accessibility issues, are well-served by Fine-tuned GPT-2. In these scenarios, the cost of data collection, model development and training can be outweighed by the overall benefits. 
In contrast, the central advantage of Prompted GPT-3.5 lies in its ease of development. It reduces the strong demand for data collection, extensive coding, and prolonged development and deployment phases. Simple natural language prompting coupled with API calls suffices for its effective utilization.
Therefore, \textit{Promptor} should be employed specifically in scenarios where data collection and development are costly, and the expenses of such do not justify the value.



\subsection{Limitations and Future Work}

The size of large language models can affect inference time, often rendering them slower than some rule-based systems. This becomes a drawback especially in applications demanding ultra-quick response times, such as auto-complete features. Nonetheless, numerous researchers are actively working towards enhancing the inference time of these models~\cite{sheng2023high,park2022nuqmm}.

Our future work aims to automate text prediction functionality entirely. Currently, we automate prompt creation and have partially developed a supporting keyboard. However, manual input is still required to place prompts into the test keyboard and set parameters in the configurable panel. We envision a fully automated process enabling personalized intelligent text prediction systems. This way, users can converse with the keyboard, which comprehends requests and self-updates to meet user needs.

\section{Conclusions}
In this study, we contrasted the performance of a Fine-tuned GPT-2 model with a Prompted GPT-3.5 model, uncovering a significant edge in favor of the latter. GPT-3.5's prompting mechanism not only streamlined the task, but also eliminated the necessity for data gathering and extensive code development. This analysis fundamentally juxtaposes an on-device large language model with an on-cloud counterpart. 
The virtues of Prompted GPT-3.5 transcend its superior performance, encompassing the domains of code-free development and obviating the need for data collection. These merits position it as an attractive option for intelligent text prediction endeavors.
Furthermore, we delved into a comparison between prompts curated by designers (with novice expertise in prompt engineering) and those crafted by \textit{Promptor}. Our observations indicated that self-designed prompts struggle to steer GPT-3.5 towards generating stable output in the desired format. They also struggle with clearly outlining the background and task guidelines for coherent predictions, and fail to use all of the provided information effectively. As a result, they prevent GPT-3.5 from producing predictions that closely match the ideal responses. In contrast, \textit{Promptor} creates prompts to guide GPT-3.5 to produce stable predictions that are not only coherent but also similar to the ideal responses. 

We envisage that the introduction of \textit{Promptor} holds potential not only in the domain of intelligent text prediction, but also the process underlying its creation, termed as \textit{Prompt2Prompt}, can provide guidelines for developers in other fields aiming to devise automated prompt generation agents.


\bibliographystyle{abbrv-doi}

\bibliography{template}
\end{document}